\documentclass[sn-apacite,Numbered]{sn-jnl}

\usepackage{microtype}
\usepackage{natbib}
\usepackage{graphicx}
\usepackage{subfigure}
\usepackage{subfiles}
\usepackage{booktabs}
\usepackage{amsmath, amssymb, mathtools, amsthm}
\usepackage[capitalize,noabbrev]{cleveref}
\usepackage{shortcuts}
\usepackage{bbold}
\usepackage{mathtools}
\usepackage{float}
\usepackage{hyperref}

\usepackage{graphicx}%
\usepackage{multirow}%
\usepackage{amsmath,amssymb,amsfonts}%
\usepackage{amsthm}%
\usepackage{mathrsfs}%
\usepackage[title]{appendix}%
\usepackage{xcolor}%
\usepackage{textcomp}%
\usepackage{manyfoot}%
\usepackage{booktabs}%
\usepackage{algorithm}%
\usepackage{algorithmicx}%
\usepackage{algpseudocode}%
\usepackage{listings}%

\usepackage{todonotes}

\newtheorem{assum}{Assumption}

\usepackage{xr}
\usepackage{xr-hyper}
\makeatletter
\newcommand*{\addFileDependency}[1]{
  \typeout{(#1)}
  \@addtofilelist{#1}
  \IfFileExists{#1}{}{\typeout{No file #1.}}
}
\makeatother

\begin{document}

\title{Exact and Approximate Conformal Inference for Multi-Output Regression}



\abstract{
  It is common in machine learning to estimate a response $y$ given covariate information $x$. However, these predictions alone do not quantify any uncertainty associated with said predictions. One way to overcome this deficiency is with conformal inference methods, which construct a set containing the unobserved response $y$ with a prescribed probability. Unfortunately, even with a one-dimensional response, conformal inference is computationally expensive despite recent encouraging advances. In this paper, we explore multi-output regression, delivering exact derivations of conformal inference $p$-values when the predictive model can be described as a linear function of $y$. Additionally, we propose \texttt{unionCP} and a multivariate extension of \texttt{rootCP} as efficient ways of approximating the conformal prediction region for a wide array of multi-output predictors, both linear and nonlinear, while preserving computational advantages. We also provide both theoretical and empirical evidence of the effectiveness of these methods using both real-world and simulated data.\looseness=-1
}








\author*[1]{\fnm{Chancellor} \sur{Johnstone}}\email{chancellor.johnstone.1@us.af.mil}

\author[2]{\fnm{Eugene} \sur{Ndiaye}}\email{e\_ndiaye@apple.com}

\affil*[1]{\orgdiv{Department of Mathematics and Statistics}, \orgname{Air Force Institute of Technology}, \orgaddress{\city{Dayton}, \state{Ohio}, \country{United States}}}

\affil[2]{\orgname{Georgia Institute of Technology, currently at Apple}}



\newcommand{\fix}{\marginpar{FIX}}
\newcommand{\new}{\marginpar{NEW}}


\maketitle
%

%






\section{Introduction}
\label{sec:intro}

In regression, we aim to predict (or estimate) some response $y$ given covariate information $x$. These predictions alone deliver no information related to the uncertainty associated with the unobserved response, and thus, would benefit from the inclusion of a set $\Gamma^{(\alpha)}(x)$ such that, for any significance level $\alpha \in (0,1)$,
\begin{equation}
    \mathbb{P}\big(y \in \Gamma^{(\alpha)}(x)\big) = 1 - \alpha.
\end{equation}
One method to generate $\Gamma^{(\alpha)}$ is through conformal inference (used interchangeably with ``conformal prediction'' in this work) \citep{gammerman1998learning, lei2018distribution}, which generates \textit{conservative} prediction sets for some unobserved response $y$ under only the assumption of exchangeability. Given a finite number of observations $\mathcal{D}_n = \{(x_i, y_i)\}_{i = 1}^n$ and a new unlabelled example $x_{n+1}$, conformal prediction regions are generated through the repeated inversion of the test, \looseness=-1
%
\begin{equation}
H_0: y_{n+1} = z  \; \textrm{ vs. } \; H_a: y_{n+1} \ne z ,
\label{eqn:conf-permute}
\end{equation}
where $z$ is a potential candidate response value, \ie the null hypothesis \citep{lei2018distribution}. A $p$-value for \Cref{eqn:conf-permute} is constructed by learning a predictive model $\hat y(z)$ on the augmented dataset $\mathcal{D}_n \cup \{(x_{n+1}, z)\}$ and comparing one's ability to predict the new candidate $z$ using $\hat y_{n+1}(z)$ to the already observed responses using, say, $\hat{y}_i(z)$, the predicted value for the $i$-th response as a function of $z$. We note that while $\hat{y}_i(z)$ depends on $D_n$, $x_i$, $x_{n+1}$, and $z$, we only explicitly highlight dependence on $z$. The so-called conformal prediction set is the collection of candidates $z$ for which the null hypothesis is not rejected, \ie when the error in predicting $z$ is not too high compared to others.

The inversion of the test is \Cref{eqn:conf-permute} is traditionally called ``full'' conformal prediction since it uses the entire dataset to learn a predictive model. Unfortunately, full conformal prediction is computationally demanding in most cases, with each new candidate point $z$ requiring a new model to be fit. To avoid this complexity, more efficient methods, \eg split conformal inference  \citep{vovk2005algorithmic, lei2018distribution} and trimmed conformal inference \citep{chen2016trimmed}, have been introduced with trade-offs between computational efficiency and performance. 

Of interest to our work in this paper are \textit{exact} and \textit{approximate conformal inference} methods, which aim to reduce computational complexity without sacrificing performance. \cite{nouretdinov2001ridge} showed that with certain models, ridge regressors in particular, conformity scores for every observation in a dataset can be constructed as an affine function of the candidate value $z$ and only require training the model once. 

%

We extend the result of \cite{nouretdinov2001ridge} to predictors of the form
\begin{equation}
    \hat{y} = Hy,
\label{eqn:colspace}
\end{equation}
%
where $y$ is an $n \times 1$ vector of responses, $H$ is an $n \times n$ matrix, and $\hat{y}$ is an $n \times 1$ vector of predictions. We note that $H$ can also be a function of a set of covariates, \eg as with ridge regression where $H = X(X^\top X + \lambda I)^{-1}X^\top$. In reality, the restriction shown in \Cref{eqn:colspace} is more general than ridge regression; we only require the predictions be linear functions of the input. In this paper, we refer to models that follow \Cref{eqn:colspace} as \textit{linear} models; this is in contrast to the traditional usage of the term to reflect models that are linear with respect to their parameters.

There also exist methods for the efficient construction of a conformal prediction set through the use of root-finding procedures. As one example of a root-finding approach for the construction of conformal prediction sets, \cite{ndiaye2021root} introduce the \texttt{rootCP} algorithm, which utilizes a traditional root-finding approach, \ie a bisection search, to find points on the boundary of a conformal prediction set with fewer model trainings that full conformal prediction. 


In more complex settings it might be of interest to construct a model for multiple responses, \ie for some response $y$ such that the support $\mathcal{Y} \subset \mathbb{R}^q$, also known as multi-output (or multi-task) regression \citep{zhang2018overview, borchani2015survey, xu2019survey}. Thus, we might wish to construct a prediction set such that some some $q$-dimension version of $y$, say $y = (y^{(1)}, \hdots, y^{(q)})^\top$, is contained with some specified probability.\looseness=-1


\paragraph{Contributions}
With these potential scenarios is mind, we aim to extend exact and approximate conformal inference to the multi-output regression setting. Specifically, we contribute:

\begin{itemize}
    \item an extension of exact conformal inference, \ie exact conformal $p$-values, to multiple dimensions with various predictors and conformity measures
    \item \texttt{unionCP} to approximate conformal prediction sets without model retraining
    \item a multivariate extension of \texttt{rootCP}
\end{itemize}  

\begin{figure}[h]
    \centering
    \resizebox{\textwidth}{!}{
    \includegraphics[trim = 25 0 70 50, clip]{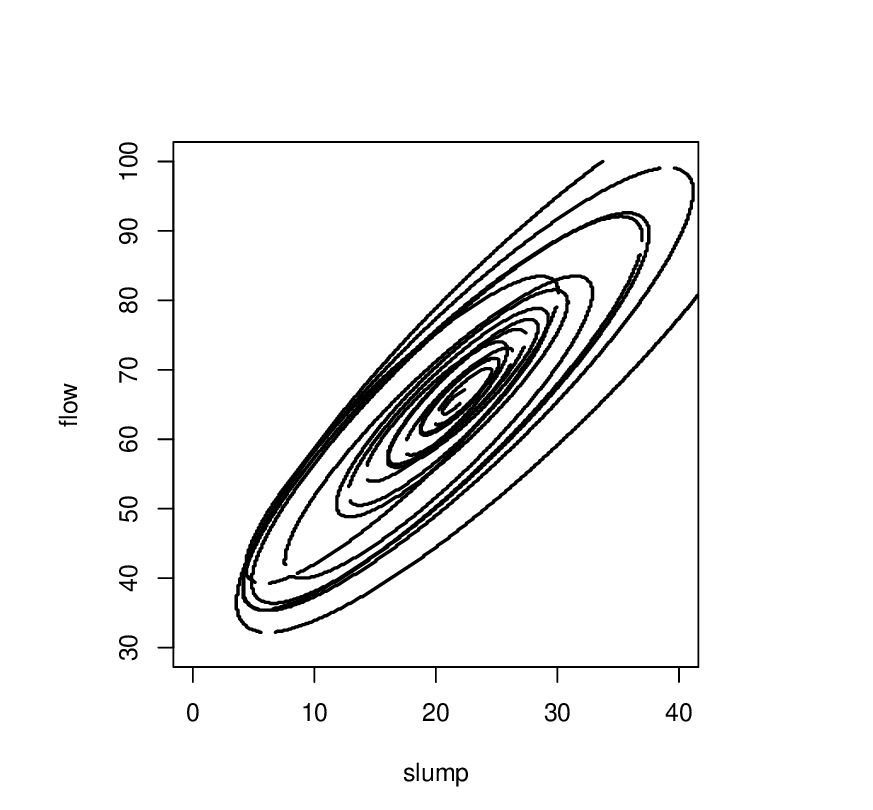}
    \includegraphics[trim = 25 0 70 50, clip]{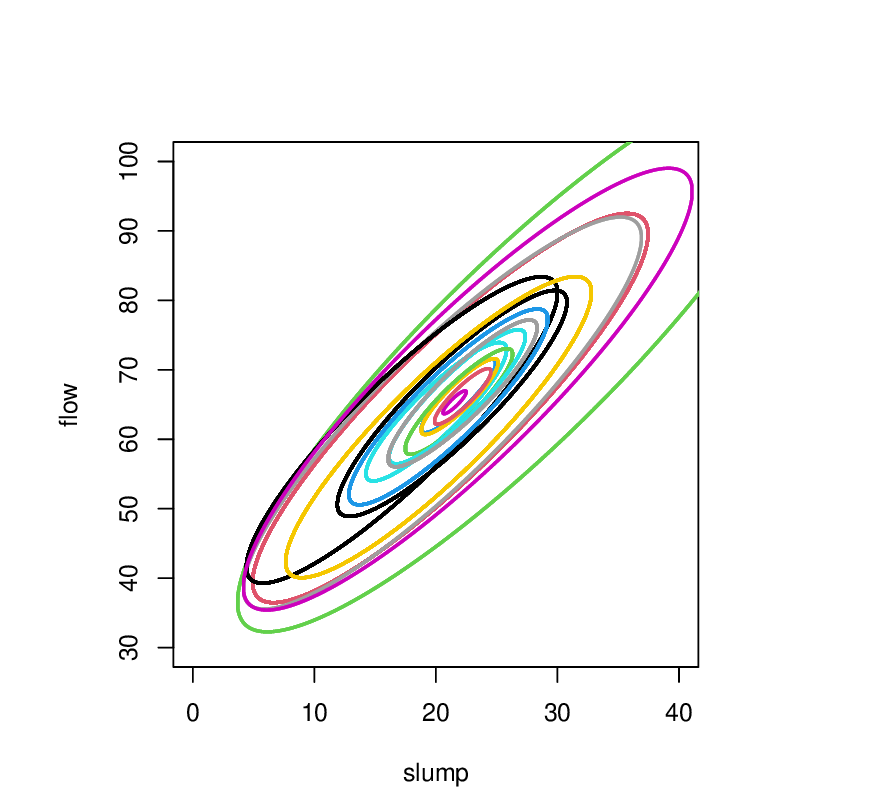}
    \includegraphics[trim = 25 0 70 50, clip]{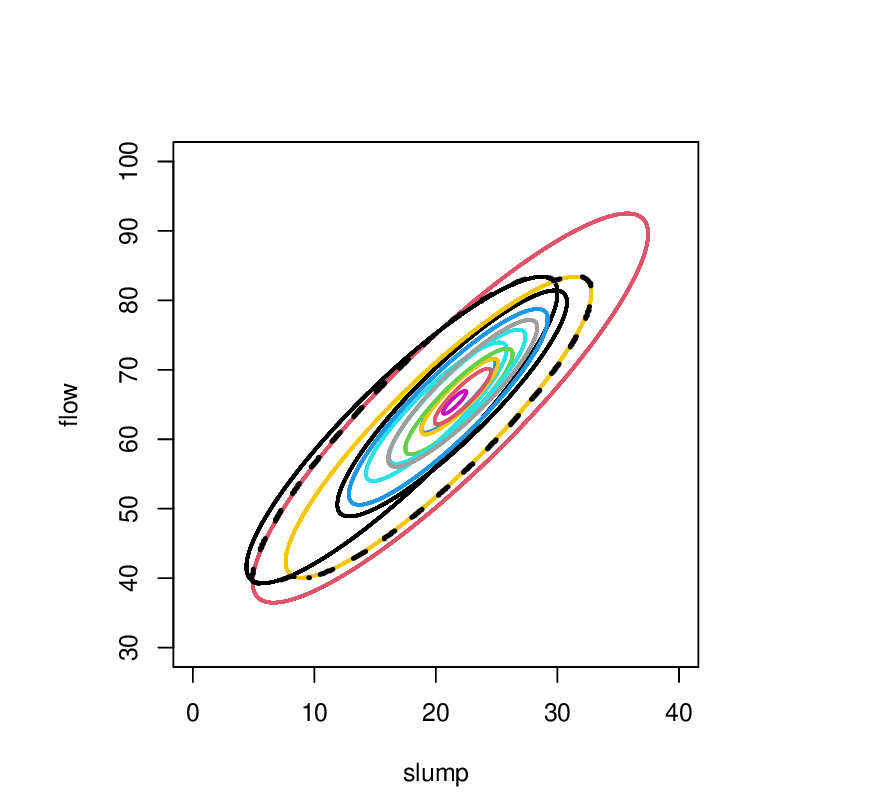}
    }
     \caption{Comparing \texttt{gridCP} contours (left) to $p$-value change-point sets (middle) constructed using $||\cdot||^2_W$ with $W = \hat{\Sigma}^{-1}$ for an observation from \texttt{cement} dataset. We also include a comparison of a \texttt{gridCP} prediction set (shown with the black line) and \texttt{unionCP} set (right) for $\alpha = 0.25$.}
    \label{fig:compare-cp}
\end{figure} 

The extension of exact conformal inference to multiple dimensions makes use of \textit{p-value change-point sets} (described in detail in Section \ref{sec:exactCP}) which allow for the description of conformal prediction regions. The introduction of \verb|unionCP| and extension of \verb|rootCP| reduce the trade-off between various conformal inference methods, balancing the computational efficiency of split conformal prediction (\verb|splitCP|) with the performance of full conformal prediction (\verb|fullCP|). Figure \ref{fig:compare-cp} compares predictive regions generated using a grid-based approach (\texttt{gridCP}) to those constructed using \texttt{unionCP}.

Thus, we can make full conformal prediction more feasible through efficient construction, exact or otherwise. Table \ref{tab:comp} summarizes the overall computational costs for each of these methods in terms of the number of model retraining iterations required to generate the conformal prediction region. We also include the computation complexity of the CP approximation provided by \texttt{gridCP}. In contrast, \texttt{fullCP} comprises approaches where the exact conformal prediction set can be constructed in a closed-form.

\begin{table}[h]
\centering
\caption{Computational complexity of methods where $q$ is the response dimension, $n$ is the number of observations, $r$ is the cardinality of the candidate value set, $d$ is the number of search directions, and $\epsilon$ is the tolerance.\looseness=-1}
\begin{tabular}{|c|c|c|}
\hline
Method             & Linear    & Nonlinear        \\
\hline
\texttt{splitCP} & $\mathcal{O}(q)$ &  $\mathcal{O}(q)$               \\
\texttt{gridCP} & $\mathcal{O}(q)$ &  $\mathcal{O}(rq)$                         \\
\texttt{fullCP}  & $\mathcal{O}(q)$ & -                  \\
\texttt{unionCP} & $\mathcal{O}(q)$ & $\mathcal{O}\big(ndq\log_2(1/\epsilon)\big)$             \\
\texttt{rootCP} & $\mathcal{O}(q)$ & $\mathcal{O}\big(dq\log_2(1/\epsilon)\big)$  \\
\hline
\end{tabular}
\label{tab:comp}
\end{table}

\noindent
From Table \ref{tab:comp}, we can see that in the linear case, each of the methods for prediction set generation require the same number of model refits as \texttt{splitCP}. We note that this does not account for the complexity of interval construction in each case.

The rest of the paper is laid out as follows. Section \ref{sec:background} provides requisite background for the paper. Section \ref{sec:exactCP} extends exact conformal inference to multiple dimensions, while Section \ref{sec:approx} introduces various conformal prediction set approximation methods in multiple dimensions. Section \ref{sec:application} provides empirical evaluation of our proposed approaches. Section \ref{sec:conclusion} concludes the paper.

\paragraph{Notation}
We denote the design matrix $X = (x_1, \ldots, x_n, x_{n+1})^\top$.
Given $j \in [n]$, the rank of an element $u_j$ among a sequence $\{u_1, \ldots, u_n\}$ is defined as
$\mathrm{Rank}(u_j) = \sum_{i=1}^{n}\mathbb{1}_{u_i \leq u_j} \enspace.$


\section{Conformal Inference}
\label{sec:background}

Originally introduced in \cite{gammerman1998learning} as ``transductive inference'', conformal inference (CI) was originally focused on providing inference with classification approaches. \cite{vovk2005algorithmic} provides a formalized introduction to conformal inference  within regression. With the express purpose of inference, the goal of CI is to attach, in some fashion, a measure of uncertainty to a predictor, specifically through the construction of a conservative prediction set, \ie one such that 
\begin{equation}
    \mathbb{P}\big(y_{n+1} \in \Gamma^{(\alpha)}(x_{n+1})\big) \ge 1 - \alpha .
    \label{eqn:conservative}
\end{equation}
We define $\mathcal{D}_n = \{(x_i, y_i)\}_{i = 1}^n$ as a collection of $n$ observations, where the $i$-th data tuple $(x_i, y_i)$ is made up of a covariate vector $x_i$ and a response $y_i$. We wish to construct a prediction set for a new observation $(x_{n+1},y_{n+1})$, where $x_{n+1}$ is some known covariate vector and $y_{n+1}$ is some, yet-to-be-observed response. Assuming each data pair $(x_i, y_i)$ and $(x_{n+1}, y_{n+1})$ are drawn exchangeably from some distribution $\mathcal{P}$, conformal inference  generates conservative, finite sample-valid prediction sets in a distribution-free manner. 

In a prediction setting, test inversion for a particular candidate value $z$ is achieved by training the model of interest on an augmented data set $\mathcal{D}_{n+1}(z) = \mathcal{D}_n \cup \{(x_{n+1}, z)\}$. At this point, we leave our model of interest general, denoting the prediction of the $i$-th observation based on a model trained with $\mathcal{D}_{n+1}(z)$ as $\hat{y}_i(z)$. 
Following the refitting, each observation in the augmented data set receives a (non)conformity \textit{measure}, which determines the level of (non)conformity between itself and other observations. One popular, and particularly effective, conformity measure is the absolute residual
\begin{equation}
\label{eqn:single-conf}
    S_i(z) = |y_i - \hat{y}_i(z)|.
\end{equation}
We can construct the conformity \textit{score} associated with a  candidate point $z$ with
\begin{equation}
\pi(z) = \frac{1}{n+1} + \frac{1}{n+1} \sum_{i = 1}^n \mathbb{1}_{S_i(z) \le S_{n+1}(z)},
\label{eqn:conf-p-values}
\end{equation}
where $S_i(z)$ is the conformity measure for the data pair $(x_i, y_i)$ as a function of $z$ and $S_{n+1}(z)$ is the conformity measure associated with $(x_{n+1}, z)$. Then, a $p$-value for the test shown in \Cref{eqn:conf-permute} can be constructed as  $1 - \pi(z)$. In Section \ref{sec:exactCP} we explicitly describe the $p$-value associated with some $z$ in terms of \textit{p-value change-point sets}, which explicitly define where changes in rank occur between the conformity scores for specific observations. 
A prediction set for an unknown response $y_{n+1}$ associated with some covariate vector $x_{n+1}$ is \looseness=-1
\begin{equation}
\Gamma^{(\alpha)}(x_{n+1}) = \{z : (n+1) \pi(z) \le \lceil(1 - \alpha)(n+1) \rceil \},
 \label{eqn:conf-pi}
\end{equation}
%
%
Then, subuniformity holds for 
$(n+1)\pi(y_{n+1}) = \mathrm{Rank}(S_{n+1}(y_{n+1})),$ and \Cref{eqn:conservative} holds for $\Gamma^{(\alpha)}(x_{n+1})$.
By the previous results, CI can also be utilized in the multivariate response case, where one is interested in quantifying uncertainty with respect to the joint behavior of a collection of responses, given a set of covariates. Thus, we can construct a multidimensional prediction set $\Gamma^{(\alpha)}(x_{n+1}) \subset \mathbb{R}^q$ such that \Cref{eqn:conservative} holds when $y_{n+1}$ is some $q$-dimensional random vector.






Previous results extending conformal inference to the multivariate setting comes from \cite{lei2015conformal}, which applies conformal inference  to functional data, providing bounds associated with prediction ``bands''. \cite{diquigiovanni2022conformal} extends and generalizes additional results for conformal inference  on functional data. Joint conformal prediction sets outside the functional data setting are explored in \cite{kuleshov2018conformal} and \cite{neeven2018conformal}. \cite{messoudi2020conformal, messoudi2021copula} provide extensions to these works through the use of Bonferroni- and copula-based conformal inference, respectively. \cite{cella2020valid}, \cite{kuchibhotla2020exchangeability} and \cite{johnstone2021conformal} construct joint conformal sets through the use of depth measures, \eg half-space and Mahalanobis depth, as the overall conformity measure. \cite{messoudi2022ellipsoidal} extends these works by generating adaptive conformal predictive regions in multiple dimensions.
Applications of conformal inference have been seen in healthcare \citep{olsson2022estimating}, drug discovery \citep{cortes2019concepts, eklund2015application, alvarsson2021predicting}, and decision support \citep{wasilefsky2023responsible}, to name a few. For a thorough treatment on conformal inference in general, we point the interested reader to \cite{fontana2023conformal} and \cite{angelopoulos2023conformal}. We also point the reader to \cite{hallin2010multivariate} for an approach to generate quantiles for multi-output regression.

\subsection{Computationally Efficient Conformal Inference }

Due to the inherent model refitting required to generate prediction sets through full conformal inference, \ie the testing of an infinite amount candidate points, more computationally efficient methods have been explored. We describe a subset of these methods in the following sections. Specifically, we focus on resampling-based and exact conformal inference.\looseness=-1

\paragraph{Resampling Methods}

Split conformal inference  \citep{vovk2005algorithmic, lei2018distribution} generates conservative prediction intervals under the same assumptions of exchangeability as \textit{full} conformal inference . However, instead of refitting a model for each new candidate value, split conformal inference  utilizes a randomly selected partition of $\mathcal{D}_n$, which includes a training set $\mathcal{I}_1$ and a calibration set $\mathcal{I}_2$. First, a prediction model is fit using $\mathcal{I}_1$. Then, conformity measures are generated using out-of-sample predictions for observations in $\mathcal{I}_2$. The split conformal prediction interval for an incoming $(x_{n+1},y_{n+1})$, when using the absolute residual as our comformity measure, is
\begin{equation}
\label{eqn:split-cp-simp}
\Gamma_{\texttt{split}}^{(\alpha)}(x_{n+1}) = [\hat{y}_{n+1} - s, \hat{y}_{n+1} + s] ,
\end{equation}
where $\hat{y}_{n+1}$ is the prediction for $y_{n+1}$ generated using the observations in $\mathcal{I}_1$, and $s$ is the $\lceil(|\mathcal{I}_2| + 1)(1-\alpha)\rceil$-th largest conformity measure for observations in $\mathcal{I}_2$.
In order to combat the larger widths and high variance associated with split conformal intervals, cross-validation (CV) approaches to conformal inference  have also been implemented. The first CV approach was introduced in \cite{vovk2015cross} as cross-conformal inference with the goal to ``smooth'' inductive conformity scores across multiple folds. Aggregated conformal predictors \cite{carlsson2014aggregated} generalize cross-conformal predictors, constructing prediction intervals through any exhangeable resampling method, \eg bootstrap resampling. Other resampling-based conformal predictors also include CV+ and jackknife+ \citep{barber2021predictive}.
For a more detailed review and empirical comparison of resampling-based conformal inference methods, we point the interested reader to \cite{contarino2022conformal}. \looseness=-1

\paragraph{Exact Conformal Inference for Piecewise Linear Estimators}

In order to test a particular set of candidate values for inclusion in $\Gamma^{(\alpha)}(x_{n+1})$, we must compare the conformity measure associated with our candidate data point to the conformity measures of our training data. Naively, this requires the refitting of our model for each new candidate value. However, \cite{nouretdinov2001ridge} showed that $S_i(z)$, constructed using \Cref{eqn:single-conf} in conjunction with a ridge regressor, varies piecewise-linearly as a function of the candidate value $z$, eliminating the need to test a dense set of candidate points through model refitting. Other exact conformal inference methods include conformal inference through homotopy \citep{lei2019fast, ndiaye2019computing}, influence functions \citep{bhatt2021fast, cherubin2021exact}, and root-finding approaches \citep{ndiaye2021root}. While not exact, \cite{ndiaye2022stable} provide approximations to the full conformal prediction region through stability-based approaches.

\section{Exact Conformal Inference for Multi-Output Regression}
\label{sec:exactCP}

In the following sections, we extend the results in \cite{nouretdinov2001ridge} to multiple dimensions. We also discuss closed-form solutions for more general predictors as well as higher dimension prediction sets with other conformity measures. While CI can be applied to any prediction or classification task, in this section we restrict each of our predictors, given an incoming observation $(x_{n+1},z)$, to the form
\begin{equation}
    \hat{y}^{(k)}(z_k) = H_k(x_{n+1}, x_i)y^{(k)}(z_k),
    \label{eqn:colspace-z}
\end{equation}
%
%
where $\hat{y}^{(k)}(z_k)$ is the vector of predictions for the $k$-th response as a function of the candidate value $z_k$, and the candidate value \textit{vector} is defined as $z = (z_1, \hdots, z_q)^\top$. We note that the restriction shown in \Cref{eqn:colspace-z} is analogous to the restriction identifed in \Cref{eqn:colspace}. Additionally, we require that $H_k$  be constructed independently of $y^{(k)}$, \ie not as a function of $y^{(k)}$. Even with this restriction, $H_k$ is general enough so as to include many classes of predictors with examples described below. As an example, we can describe $H_k$  with respect to the $k$-th response dimension for ridge regression as (see the supplementary materials for more examples) \looseness=-1
\begin{equation}
    H_k(x_{n+1}) = X(X^\top X + \lambda_k I)^{-1}X^\top.
    \label{eqn:q-mat}
\end{equation}
%
One focus of our paper is construction of exact $p$-values for a given $z$ without retraining our model. We also identify how we construct explicit \textit{p-value change-point sets}, denoted as $\mathcal{E}_i$ for the $i$-th observation, where 
\begin{equation}
    \mathcal{E}_i \equiv \{z \in \mathbb{R}^q: S_{n+1}(z) \le S_{i}(z)\},
    \label{eqn:pval-change}
\end{equation} 
with the end goal of generating exact conformal prediction sets. We note that $\mathcal{E}_{n+1} \equiv \mathbb{R}^q$. Then, the $p$-value associated with the hypothesis test shown in \Cref{eqn:conf-permute} for any candidate point $z$ is 
\begin{equation}
    \textrm{$p$-value}(z) = \frac{|\{i \in [n+1]: z \in \mathcal{E}_i\}|}{n+1} .
    \label{eqn:pval}
\end{equation}



The result of \cite{nouretdinov2001ridge} was extended to include both lasso and elastic net regressors in \cite{lei2019fast}. For this paper, we utilize a generalized version, shown in Proposition \ref{thm:nouretdinov-general}.

\begin{proposition}
\label{thm:nouretdinov-general}
Assume the fitted model as in \Cref{eqn:colspace}, where $H(x_{n+1}, x_i) = H$. Then, if we define $y(z) = (y^\top,z)^\top$, we can describe the vector of residuals associated with the augmented dataset and some candidate value $z$ as 
$\hat{y}(z) - Hy(z) = A-Bz$
\noindent
where $A = \big(I - H\big)y(0)$ and $B = \big(I - H\big)(0,\hdots,0,1) ^\top$.
%
%
%

\end{proposition}




\noindent
With Proposition \ref{thm:nouretdinov-general}, we can then describe the conformity measure for the $i$-th observation, when using \Cref{eqn:single-conf}, as $S_i(z)  = |a_i + b_iz|$.
In the following sections, we describe exact conformal inference results for two conformity measure constructions, $\ell_1$ and $||\cdot||^2_W$, as well as results for finding points on the boundary of a conformal prediction set for any conformity measure.

\subsection{Exact $p$-values with $\ell_1$}
\label{sec:conf-l1} 

We formalize our extension of \cite{nouretdinov2001ridge} to multiple dimensions, specifically utilizing
%
%
\begin{equation}
\label{eqn:multi-conf-ridge}
    S_i(z) = ||y_{i} - \hat{y}_{i}(z)||_1 ,
\end{equation}
as our conformity measure, in Proposition \ref{thm:l1}.

\begin{proposition}
\label{thm:l1}
Assume the fitted model, $\hat{y}^{(k)}(z_k) = H_k(x_{n+1},x_i)y^{(k)}(z_k)$. Then, using $\Cref{eqn:multi-conf-ridge}$,
$S_i(z) = ||a_i + b_iz||_1,$
where $a_i = (a_{1i}, \hdots, a_{qi})^\top$, $b_i = (b_{1i}, \hdots, b_{qi})^\top$, and $a_{ki}$ and $b_{ki}$ are the $i$-th elements of the vectors $A_k$ and $B_k$, respectively, defined as
%
\begin{equation}
    \label{eqn:ak_bk} 
    A_k = \big(I - H_k(x_{n+1}, x_i) \big)y^{(k)}(0) \text{ and } B_k = \big(I - H_k(x_{n+1}, x_i) \big)(0,\hdots,0,1)^\top. 
\end{equation} 

\end{proposition}


Proposition \ref{thm:l1} allows us to construct conformity measures associated with a multidimensional response without retraining the model for each new $z$. Additionally, using Proposition \ref{thm:l1} for each observation $(x_i, y_i)$, we can generate a region $\mathcal{E}_i$, as defined in \Cref{eqn:pval-change}. We can also construct a fixed-point solution for 
$\hat{y}_{n+1}(z)$, \ie a point where $\hat{y}_{n+1}(z) = z$, as
\begin{equation}
    \tilde{z} = \bigg(-\frac{a_{1n+1}}{b_{1n+1}}, \hdots, -\frac{a_{qn+1}}{b_{qn+1}} \bigg).
    \label{eqn:tildez-q}
\end{equation}
\noindent

\noindent
\Cref{eqn:tildez-q} can be derived by setting each component of $S_{n+1}(z)$ equal to zero; the fixed point for a given observation is where the probability of a more extreme response, \ie $p\textrm{-value}(z)$, is maximized. 

It is initially unclear \textit{how} the construction of an individual region $\mathcal{E}_i$ occurs when using $\ell_1$.  As it stands, finding all $z$ such that $S_i(z) = S_{n+1}(z)$ is a multidimensional root-finding problem with infinite solutions, which has exponential complexity as $q$ increases. However, we have seen empirically that $\mathcal{E}_i$ constructed with $\ell_1$ exhibits consistent structure. Namely, when using $\ell_1$, each $\mathcal{E}_i$ can be defined by the convex hull of a collection of points, specifically points axis-aligned with the fixed-point solution $\tilde{z}$. These points, referred to as ``corners'' within this paper, differ from $\tilde{z}$ in only the $j$-th element. 


In order to provide clarity, we include Algorithm 1 to construct $\mathcal{E}_i$ in practice when using $\ell_1$ as well as a two-dimensional visual of the solutions generated for an observation from the \verb|cement| dataset \citep{yeh2007modeling} in Figure \ref{fig:quick}. We also include further discussion on Algorithm 1 in Supplementary Materials \looseness=-1 

\begin{figure}[h]
    \centering
    \resizebox{.7\textwidth}{!}{
    \includegraphics[trim = 25 0 50 50, clip]{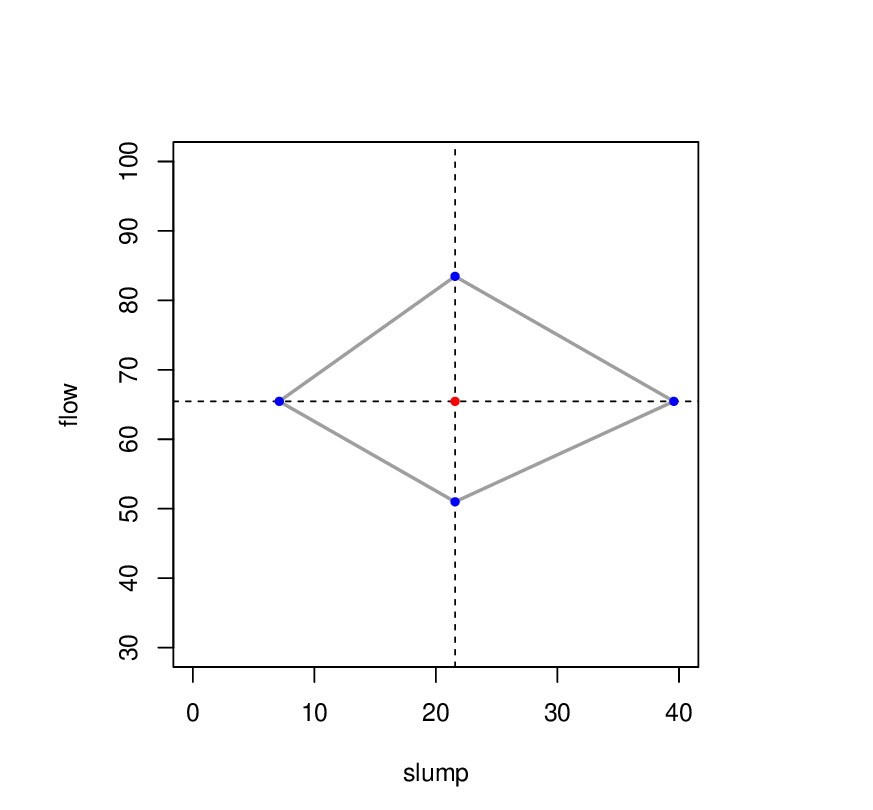}
    \includegraphics[trim = 25 0 50 50, clip]{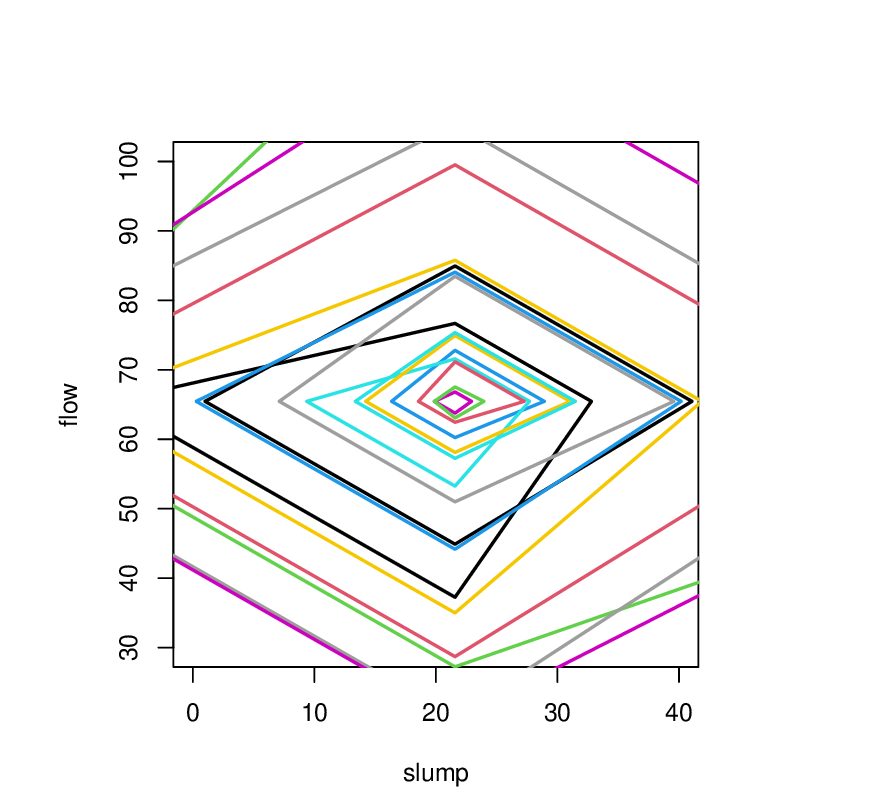}
    }
    \caption{Example of Algorithm 1 for constructing $\mathcal{E}_i$ for an observation from \texttt{cement} dataset (left). The ``\textcolor{red}{$\bullet$}'' identifies $\tilde{z}$, while the grey line represents the border of the $p$-value change-point set. Corner points are identified with ``\textcolor{blue}{$\bullet$}''. The axes generated with $\tilde{z}$ are shown with the dotted black lines. We also include the collection of $p$-value change-point sets (right).}
    \label{fig:quick}
\end{figure}

\subsection{Exact $p$-values with $||\cdot||^2_W$}
\label{sec:conf-ell}

In order to generalize our exact $p$-value construction further than for use solely with $\ell_1$, we now consider conformity measures of the form
\begin{equation}
\label{eqn:multi-conf}
    S_i(z) = r_i(z)^\top Wr_i(z) = ||r_i(z)||^2_W,
\end{equation}
where $r_i(z) = y_i - \hat{y}_i(z)$, and $W$ is some $q \times q$ matrix.  Proposition \ref{thm:quad-d} provides a similar result to Proposition \ref{thm:l1}, but instead utilizes \Cref{eqn:multi-conf}. Namely, $S_i$ becomes quadratic with respect to $z$, instead of piecewise-linear. 


%
%







\begin{proposition}
\label{thm:quad-d}
Assume the fitted model $\hat{y}^{(k)}(z_k) = H_k(x_{n+1},x_i)y^{(k)}(z_k)$ for each response dimension $k \in [q]$. Then, using $\Cref{eqn:multi-conf}$,
\begin{align}
\notag S_i(z) & =   
\begin{bmatrix} 
    a_{1i} + b_{1i}z_1 \\
    \vdots \\
    a_{qi} + b_{qi}z_q  
\end{bmatrix}
^\top W
\begin{bmatrix} 
    a_{1i} + b_{1i}z_1 \\
    \vdots \\
    a_{qi} + b_{qi}z_q  
\end{bmatrix}
\end{align}
where $a_{ki}$ and $b_{ki}$ are the $i$-th elements of the vectors $A_k$ and $B_k$, respectively, as defined in \Cref{eqn:ak_bk}.

\end{proposition}


%

In order to maintain the probabalistic guarantees inherent to conformal inference, we require $W$ to be constructed exchangeably. Two constructions that satisfy exchangeability are: 1) $W$ constructed independently of $\mathcal{D}_{n+1}(z)$, or 2) $W$ constructed using all observations within $\mathcal{D}_{n+1}(z)$. However, we show in Section \ref{sec:application} that, in practice, setting $W = \hat{\Sigma}^{-1}$, the observed inverse-covariance matrix associated with the residuals from our $q$ responses using a model constructed using only $\mathcal{D}_n$, performs well. The $p$-value associated with some $z$ using sets constructed using \Cref{eqn:multi-conf} is the same as in \Cref{eqn:pval}.

While Proposition \ref{thm:quad-d} does not restrict the structure of $W$, limiting $W$ to be a symmetric, positive semi-definite matrix ensures that the set $\mathcal{E}_i$ is not only convex, but ellipsoidal. Without this additional restriction on the matrix $W$, the $p$-value change-point sets could be ill-formed, \ie non-convex. An example of an ill-formed $p$-value change-point set is shown in Supplemtary Materials. For clarity, we also include in Supplementary Materials Algorithm 2, which describes how each $\mathcal{E}_i$ can be constructed in practice when using $||\cdot||^2_W$.


\subsection{Exact Directional Conformal Prediction Set}

To cope with higher-dimensional complexity, we now aim to sample points on the boundary of the conformal prediction set. The idea is that by picking an arbitrary direction of the space, the computation of the conformal prediction set restricted to that direction is amenable to a one dimensional problem. As such, by varying the search direction, we can easily and exactly sample several points in the boundary of the conformal region and approximate it with a convex set.
We can describe our results more generally by finding roots to $S_{n+1}(z) - S_i(z)$ where $z$ belongs to a line co-linear to the direction $d$ and passing through a conformal point $z_0$, \ie points described as
\begin{equation}
    z(t,d) = z_0 + td, \quad t \in \mathbb{R}   
    \label{eqn:z}
\end{equation}
for some direction vector $d \in \bbR^q$ and some interior point $z_0$. The boundary points are $z(t^\star, d)$ where the scalars $t^\star$ are solution of
$$S_{n+1}(z(t^\star,d)) = S_i(z(t^\star,d))$$
Restricting our models to linear predictors that follow \Cref{eqn:colspace} in one direction of the space is equivalent to restricting the observed output. As such, we have
$$\hat y(z(t, d)) = H y(z_0) + t H (0, \ldots, 0, d)^\top$$
and the conformity scores along the direction $d$ are then given by
\begin{align*}
     &S_i(z(t,d)) = \norm{a_i - t b_i},
     &S_{n+1}(z(t,d)) = \norm{a_{n+1} - t b_{n+1}},
\end{align*}
for some explicit data dependent quantities $(a_i, b_i)_{i \in [n+1]}$ described in Supplementary Materials.

The goal is now to solve the one dimension problem $ \psi(t) = S_{n+1}(z(t,d)) - S_i(z(t,d)) \geq 0 $. Without this one dimensional restriction, the computations are significantly more difficult and impossible to track without stronger data assumptions. This is illustrated in Supplementary Materials where we provide simple examples that lead to a non-convex set of solutions.
For completeness, we describe in the appendix the solution for different norms to be used as score functions and explicit form of the conformal set for a given direction.

Given an arbitrary direction $d$, one can explicitly describe the set of conformal values in that direction. Indeed, for any $i$ in $[n+1]$, we denote the intersection points of the functions $S_{i}(z_{t})$ and $S_{n+1}(z_{t})$ by $t_{i}$s and then we have $\mathcal{E}_i$ can be an interval (possibly a point), a union of interval or even empty. In all cases, it is characterized by the intersection points obtained explicitly. Since $\pi(z(t,d))$ is piecewise constant, it changes only at those points.
%
%
We denote the set of solutions $t_{1}, \cdots, t_{K}$ in increasing order as $t_{0} < t_{1} < \cdots < t_{K}$. Whence for any $t$, it exists a unique index $j=\mathcal{J}(t)$ such that $t \in (t_{j}, t_{j+1})$ or $t \in \{t_{j}, t_{j+1}\}$ and for any $t$, we have \looseness=-1
$$
(n+1)  \pi(z_{t}) = \sum_{i=1}^{n+1} \mathbb{1}_{t \in S_i}
= N(\mathcal{J}(t)) + M(\mathcal{J}(t))
$$
where the functions
$$N(j) = \sum_{i=1}^{n+1} \mathbb{1}_{ (t_{j}, t_{j+1}) \subset S_i} \text{ and } M(j) = \sum_{i=1}^{n+1} \mathbb{1}_{ t_{j} \in S_i}$$
Note that $\mathcal{J}^{-1}(j) = (t_{j}, t_{j+1}) \text{ or } \mathcal{J}^{-1}(j) = \{t_{j}, t_{j+1}\}$.
The restriction of the conformal set to the direction $d$ is
\begin{align}
\Gamma^{(\alpha)}(x_{n+1}, d) &= \bigcup_{\underset{N(j)> (n+1)\alpha}{j \in [K]}}  \!\!\! (t_{j}, t_{j+1}) \;\; \cup \!\!  \bigcup_{\underset{M(j)> (n+1)\alpha}{j \in [K]}}  \{t_{j}\} \enspace.
\end{align}

\section{Approximate Conformal Inference for Multi-Output Regression}
\label{sec:approx}

While the results in Section \ref{sec:exactCP} allows for the construction of exact $p$-values with no additional model refitting (for multiple responses), we still cannot describe exactly the conformal prediction sets in closed-form. Thus, we aim to construct approximations of the conformal prediction set for a given $x_{n+1}$. 
In this section we specifically introduce a union-based approximation for a conformal prediction set generated using the results from Section \ref{sec:exactCP}. Additionally, we extend the root-based approximation procedures introduced in \cite{ndiaye2021root} to the multi-output setting.

\subsection{\texttt{unionCP} Approximation Method}
\label{sec:unionCP}

After constructing the set $\mathcal{E}$ for an incoming point $x_{n+1}$, it is initially unclear which regions $\mathcal{E}_i$ make up various conformal prediction sets, let alone how we need to combine these regions to get the exact conformal prediction sets. Thus, we aim to provide an approximation of conformal prediction sets using the regions generated with the approaches introduced in Section \ref{sec:exactCP}. We provide Proposition \ref{thm:any} to bound error probabilities associated with potential combinations of these regions.

\begin{proposition}
\label{thm:any}
Under uniqueness of conformity measures, for some $y_{n+1}$ such that $(x_1,y_1), \hdots, (x_{n+1}, y_{n+1})$ are drawn exchangeably from $\mathcal{P}$, for any $\mathcal{S} \subset [n]$, it holds
$\mathbb{P}\Big(y_{n+1} \in \bigcup_{i \in \mathcal{S}} \mathcal{E}_i \Big) \ge  \frac{|\mathcal{S}|}{n+1} .$
\end{proposition}

Proposition \ref{thm:any} states that with the selection of \textit{any} subset of $\mathcal{E}$, the probability of the response $y_{n+1}$ being contained in the union of that subset is bounded-below by a function of cardinality. For example, if we wish to construct, say, a conservative 50\% prediction set, we could select (at random) a set $\mathcal{S} \subset [n]$ such that $|\mathcal{S}| \ge |\mathcal{E}|/2$; the union of all sets within $\mathcal{S}$ would provide a conservative prediction set. We again note that while our work emphasizes $\ell_1$ and $||\cdot||^2_W$, Proposition \ref{thm:any} holds for any conformity measure.\looseness=-1

Now, the random set constructed might not provide tight coverage as there exist some $\mathcal{E}_i$ such that $\bigcup_{i' \in \mathcal{S}_{(i)}} \mathcal{E}_{i'} \subseteq \mathcal{E}_i,$ where $\mathcal{S}_{(i)}$ is some subset of $[n]$ that does not contain $i$; some $p$-value change-point sets are contained in others and, thus, choosing the larger set could result in extremely conservative coverage. We include results related to the theoretical coverage associated with the randomized approach in Supplemental Materials.\looseness=-1

While the union of a random selection of regions forms a conservative $1 - |\mathcal{S}|/(n+1)$ prediction set, we can provide more intelligently constructed sets that are empirically less conservative (but still valid). Suppose we provide an ordering of our regions, where $\mathcal{E}_{(k)}$ is defined as the $k$-th smallest region by volume. 

\begin{definition}[\texttt{unionCP}]
A smaller $(1-\alpha)$ prediction set approximation can then be constructed as 
\begin{equation}
\hat{\Gamma}^{(\alpha)}(x_{n+1}) = \bigcup_{i \in \mathcal{S}_{1-\alpha}} \mathcal{E}_{(i)},    
\label{eqn:smart}
\end{equation}
where $\mathcal{S}_{1-\alpha} = [\lceil(1-\alpha)(n+1) \rceil]$. We dub the approximation shown in \Cref{eqn:smart} as \verb|unionCP|.
\end{definition}

By Proposition \ref{thm:any}, \verb|unionCP| generates an approximation that, at minimum, provides a region that is at least valid. We compare prediction sets constructed using \verb|unionCP| to a random selection of regions for multiple predictors in Section \ref{sec:application}. We find empirically that sets constructed using \texttt{unionCP} are less conservative than a random collection of $p$-value change-point sets.

While \Cref{thm:any} and the adjustment described in \Cref{eqn:smart} allow for conservative prediction sets, at times, the union of various $\mathcal{E}_i$ does not explicitly describe a conformal prediction set exactly. Thus, \texttt{unionCP} provides (at worse) a conservative approximation of the true conformal prediction set. We also note that \texttt{unionCP} requires the computation of volumes for each change-point set. While this is not an issue when using $\ell_1$ or $||\cdot||^2_W$ as the conformity measure, it might prohibitive for other conformity measures.\looseness=-1


With full comformal prediction, the computational complexity depends heavily on the number of candidate values chosen, while the computational burden of \texttt{unionCP} depends on the number of observations $n$. To reduce the computation required to generate the approximation, we can utilize results associated with subexchangeability, a lemma for which is included in Supplementary Materials.  Namely, we can randomly select any $m$ observations, where $1 < m \le n$, and the conformity measures of this subset, along with $S_{n+1}(z)$, will also be exchangeable. Thus, we can randomly select a subset of $\mathcal{E}$ of size $m$, defined as $\mathcal{E}^m$, and then order this subset by volume, where $\mathcal{E}^m_{(k)}$ is defined as the $k$-th smallest region by volume of the set $\mathcal{E}^m$. Then, by Proposition \ref{thm:any}, \texttt{unionCP} constructed with this subset also provides valid prediction regions, at a potentially much lower computational cost.

If we wish to avoid the \texttt{unionCP} approximation, we can generate exact $p$-values using \Cref{eqn:pval} in conjunction with \texttt{gridCP} for much computational gain over that of full conformal prediction. We also explore the connection between \texttt{unionCP} and \texttt{splitCP} in Supplementary Materials.\looseness=-1


\subsection{Root-based Approximation Methods}
\label{sec:root-approx}


\begin{figure*}[t]
    \centering
    \resizebox{\textwidth}{!}{
    \subfigure[$3$ search directions]{\includegraphics[width=0.5\textwidth]{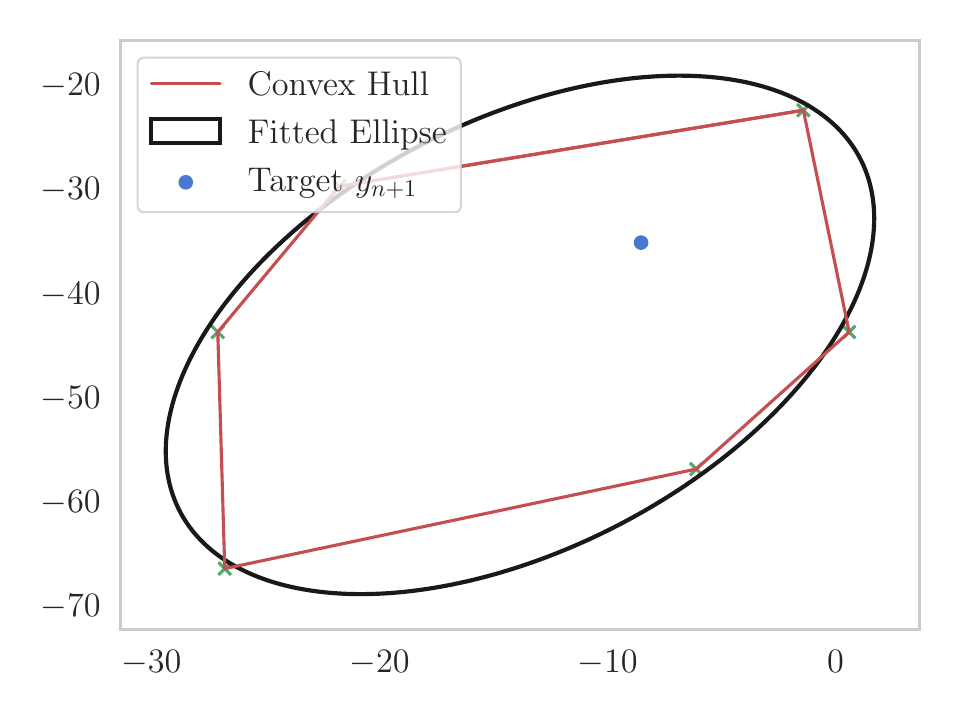}}%
    \subfigure[$5$ search directions]{\includegraphics[width=0.5\textwidth]{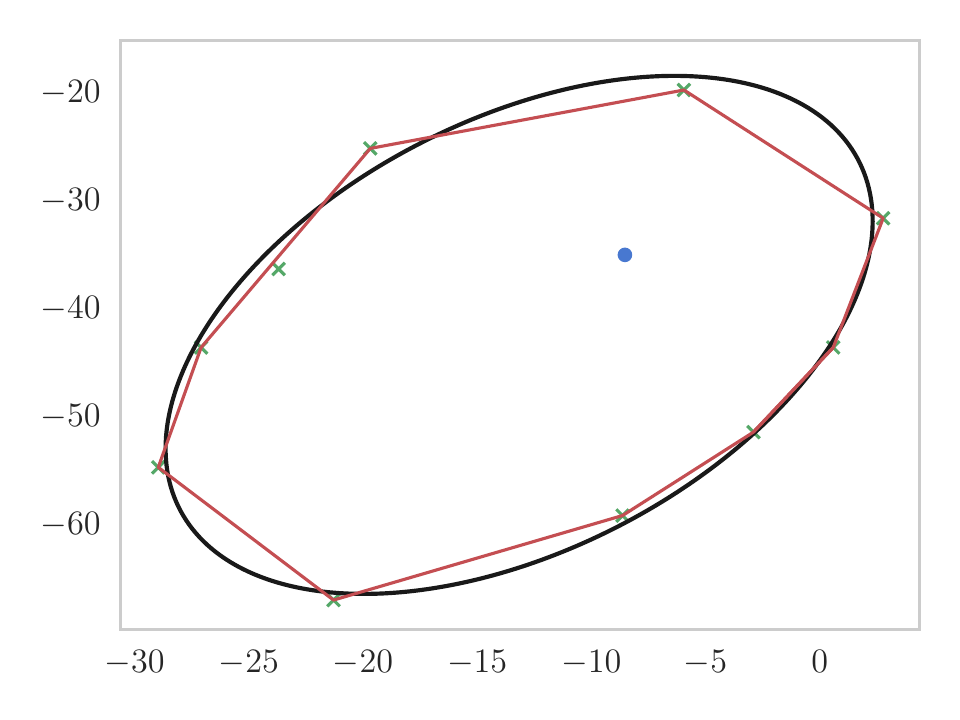}}%
    \newline
    \subfigure[$10$ search directions]{\includegraphics[width=0.5\textwidth]{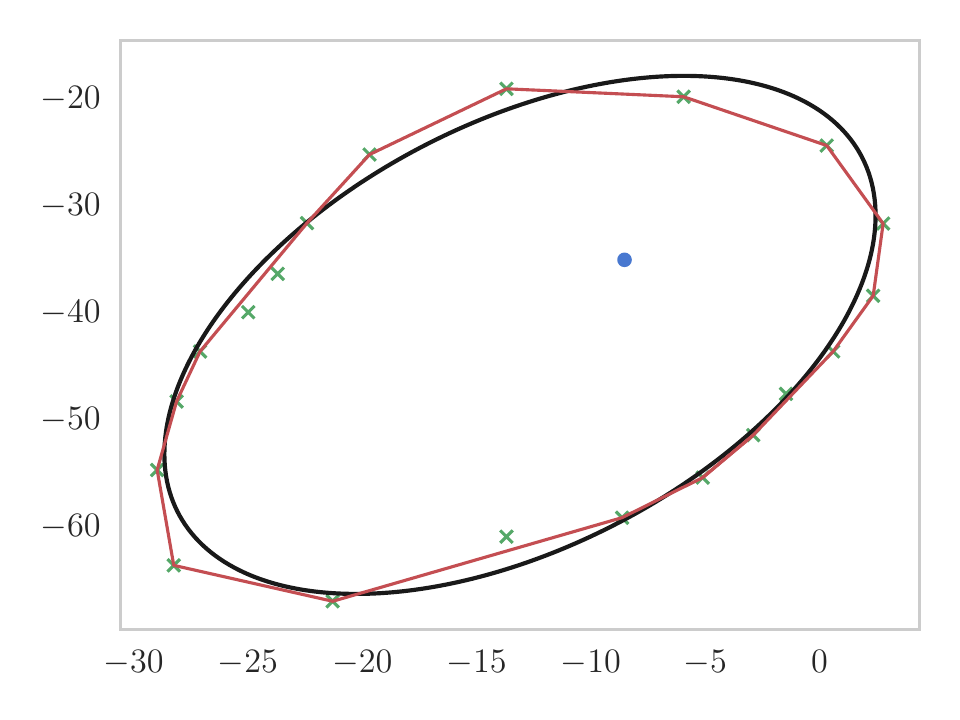}}%
    \subfigure[$30$ search directions]
    {\includegraphics[width=0.5\textwidth]{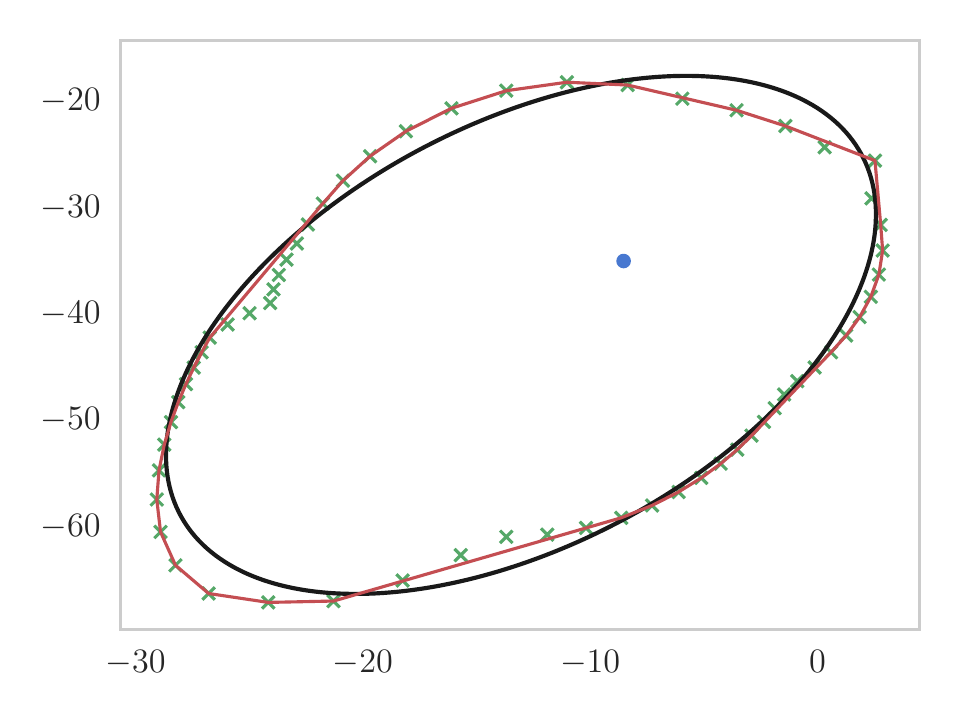}}
    }
    \caption{Illustration of the approximated conformal prediction set obtained fitting ellipse and convex hull given boundary points obtained by \texttt{rootCP}. We use scikit-learn \texttt{make\_regression} to generate synthetic dataset with the parameters
    n\_samples = $15$, n\_features = $5$, n\_targets = $2$ is the dimension of in output $y_{n+1}$. We selected $80\%$ of informative features and $60\%$ for effective rank (described as the approximate number of singular vectors required to explain most of the input data by linear combinations) and the standard deviation of the random noise is set to $5$.}
    \label{fig:root-1}
\end{figure*}

\begin{figure*}[t]
    \centering
    \resizebox{\textwidth}{!}{
    \subfigure{\includegraphics[width=0.5\textwidth]{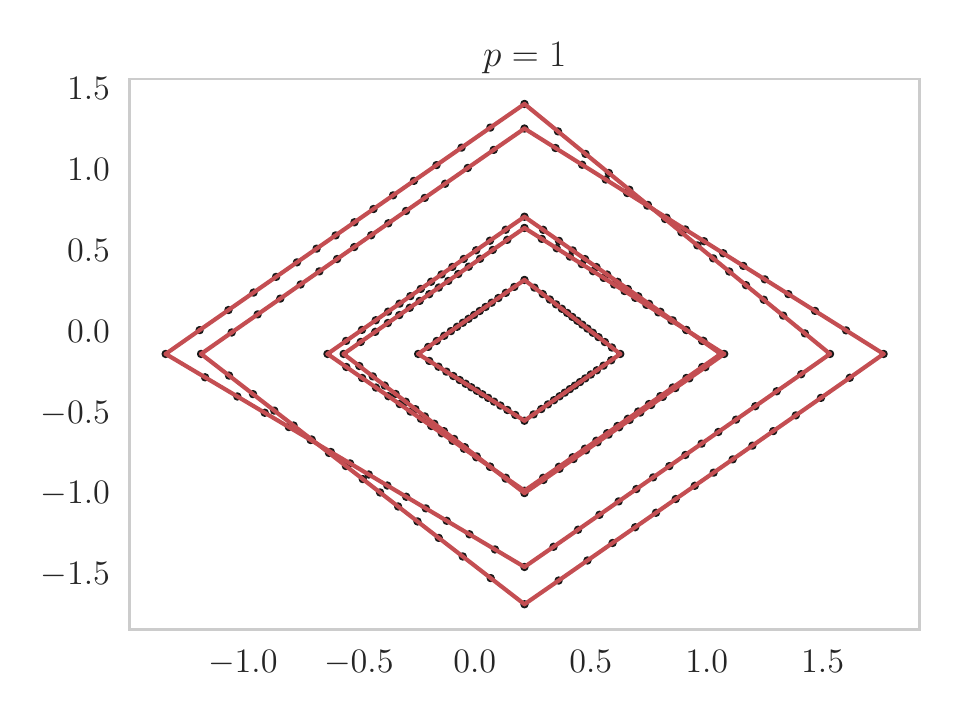}}%
    \subfigure{\includegraphics[width=0.5\textwidth]{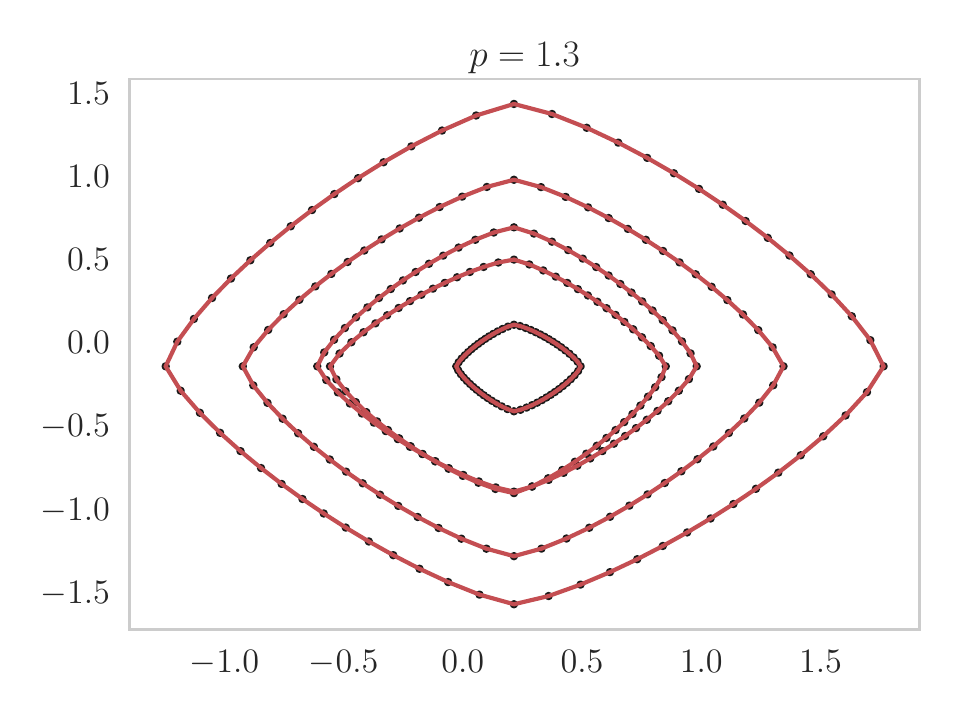}}%
    \newline
    \subfigure{\includegraphics[width=0.5\textwidth]{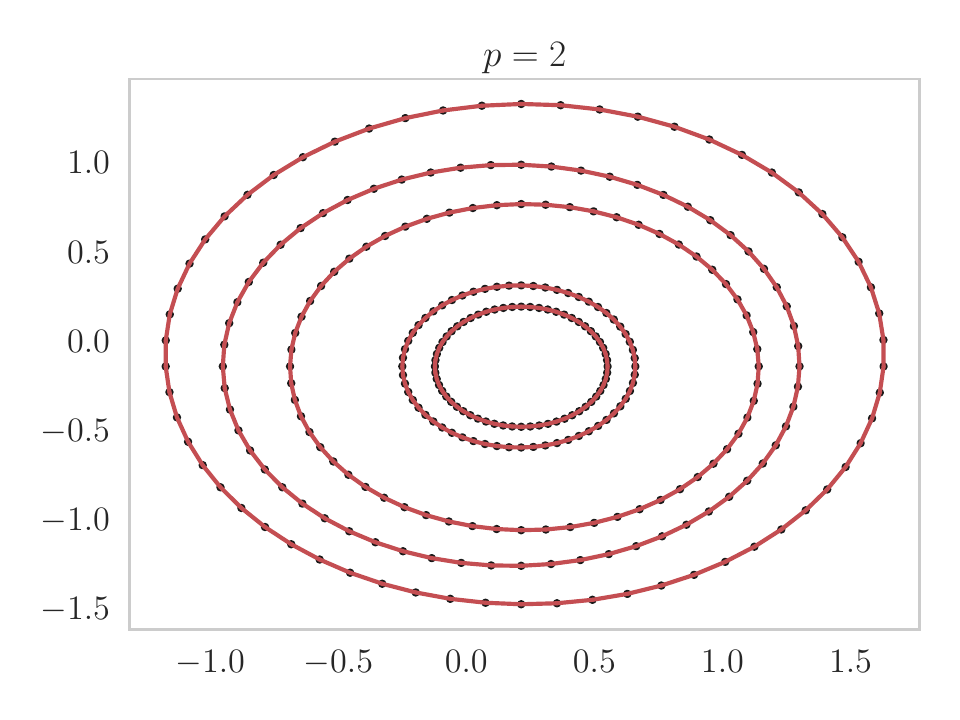}}%
    \subfigure
    {\includegraphics[width=0.5\textwidth]{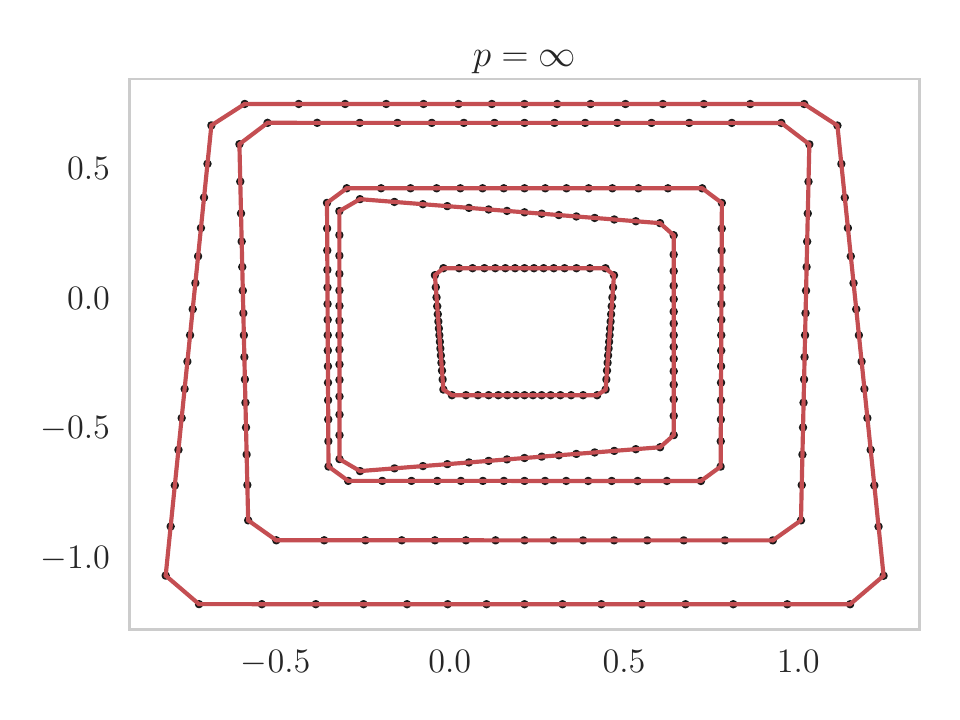}}
    }
    \caption{Illustration of the approximated $\mathcal{E}_i$ with $30$ search directions with conformity measures defined with $\ell_p$ norms. Solid black lines denote convex hull approximations of each $\mathcal{E}_i$ using calculated boundary points.}
    \label{fig:approximate_level_sets}
\end{figure*}

As noted earlier, computation of the conformal prediction sets requires model re-fitting for any candidate value to replace the true $y_{n+1}$ value. Current efficient approaches to exact computation, limited to dimension one, are restricted to models that are piecewise-linear; this structure allows to track changes in the conformity function. We have extended these approaches to higher dimensions in the previous section. 
To go beyond linear structures, we can use approximate homotopy approaches as in \citep{ndiaye2019computing} which, given an optimization tolerance, provide a discretization of all the values that $y_{n+1}$ can take. However, these approaches are also limited in dimension one and have an exponential complexity in the dimension of $y_{n+1}$. Convexity assumptions are also required, which, unfortunately, are not verified for more complex prediction models. \looseness=-1

We extend the approximations of conformal prediction in multiple dimensions by computing conformal prediction set boundaries directly. Unlike the one-dimensional case where the boundary is often two points, in multiple dimensions the boundary is continuous and, thus, uncountable, which makes finite-time computation impossible. To get around this difficulty, we will first fix a finite set of search directions; we will estimate the intersection points between the boundary of the conformal prediction set and the chosen direction. Then, we use the points on the boundary as a data base to fit a convex approximation, \eg an ellipse or the convex hull, passing through these points. This estimates the set described in \Cref{eqn:conf-pi} and is formally described below. \looseness=-1 


\begin{assum}We suppose that the conformal prediction set is \textit{star-shaped} \ie there exists a point $z_0$ such that any other point $z$ within $\Gamma^{(\alpha)}(x_{n+1})$ can be connected to $z_0$ with a line segment.
\end{assum}

A star-shaped set are not necessarily convex. We provide some illustration in \Cref{fig:root-1}. We remind that ellipsoidal sets (or any convex set) are inherently star-shaped.\looseness=-1


\paragraph{Outline of \texttt{rootCP}}
For a direction $d \in \bbR^q$, the intersection points between the boundary of $\Gamma^{(\alpha)}(x_{n+1})$
and the line passing through $z_0$ and directed by $d$ are obtained by solving the one dimensional equation 
\begin{equation}\label{eq:directional_conformity}
\pi(z(t, d)\big) = 1-\alpha, \text{ where } z(t, d) = z_0 + t d.
\end{equation}
We briefly described the main steps and display the detail in \Cref{alg:rootCP}. \looseness=-1
%
%
\begin{enumerate}
\item Fit a model $\mu_0$ on the observed training set $\Data_n$ and predict a feasible point $z_0 = \mu_0(x_{n+1})$.
\item For a collection of search directions $\{d_1, \ldots, d_K\}$, perform a bisection search in $[t_{\min}, 0]$ and $[0, t_{\max}]$ to output solutions $\hat \ell(d_k)$ and $\hat u(d_k)$ of \Cref{eq:directional_conformity} at direction $d_k$, after at most $\log_2(\frac{t_{\max} - t_{\min}}{\epsilon_r})$ iterations for an optimization tolerance $\epsilon_r >0$. Notice that the star-shape assumption implies that we will have only two roots on the selected directions.
\item Fit a convex set on the roots obtained at the previous step $\{\hat \ell(d_k), \hat u(d_k)\}_{k \in [K]}$. In practice, when one uses a least-squares ellipse as the convex approximation, a number of search directions $K$ proportional to the dimension $q$ of the target $y_{n+1}$ is sufficient.  This is not necessarily the case for the convex hull. We refer to \Cref{fig:root-1} where we observe that many more search directions are needed to cover the conformal set when using the convex hull approximation.
\end{enumerate}

The root-finding approach can also be adapted to \texttt{unionCP} by approximating the level-line boundary of the $\mathcal{E}_i$ score difference introduced in \Cref{eqn:pval-change}. In so doing, the previous restriction to quadratic functions that enabled an explicit construction is no longer necessary, at the cost of an approximation. We illustrate this generalization to different score functions in \Cref{fig:approximate_level_sets}.



    

\section{Empirical Results and Application}
\label{sec:application}

To provide empirical support for our theoretical results we consider four multi-output regression data sets. These include the data sets shown in Table \ref{tab:data-tab}. We limit the number of observations to no more than 500 for computational ease.

\begin{table}[h]
\centering
\caption{Summary of data sets used in empirical exploration.}
\begin{tabular}{|c|c|c|c|c|}
\hline
Data   & \# features & \# responses & n & Source \\
\hline
cement & 7           & 3       & 104 & \cite{yeh2007modeling} \\
\hline
enb    & 8           & 2       & 500+ & \cite{tsanas2012accurate} \\
\hline
jura   & 13          & 3       & 360 & \cite{goovaerts1997geostatistics} \\
\hline
puma32h   & 27          & 6       & 500+ & \cite{corke1996robotics} \\
\hline
\end{tabular}
\label{tab:data-tab}
\end{table}

By subuniformity, $\pi(y_{n+1}) \geq \alpha$ with probability larger than $1 - \alpha$. Hence, we can define the \texttt{oracleCP} as $\pi^{-1}([\alpha, +\infty))$ where $\pi$ is obtained with a model fit optimized on the oracle data $\Data_{n+1}(y_{n+1})$ on top of the root-based approach to find boundary points. We remind the reader that the target variable $y_{n+1}$ is not available in practice.

The following sections include results for both the application of \texttt{unionCP} and \texttt{rootCP}.

\subsection{\texttt{unionCP} Approximation Application}
\label{sec:union-app}

While our focus for the construction of $H(x_{n+1}, x_i)$ has been general, for much of our discussion we often reference $H(\cdot)$ constructions associated with a ridge-regressor. Thus, in this section, we wish to demonstrate the performance of \texttt{unionCP} with other methods that fall under our the general model restriction; we specifically explore ridge regression (RR), local-constant regression (LC) and local-linear regression (LR), each of which are discussed further in Supplementary Materials. We explore empirical coverage using each of these predictions for $\alpha = \{.1, .2, .5, .8, .9\}$. Additionally, we also construct prediction regions using the randomized approach discussed in Section \ref{sec:unionCP} for comparison.  \looseness=-1

\begin{figure*}[t]
    \centering
    \resizebox{\textwidth}{!}{
    \includegraphics{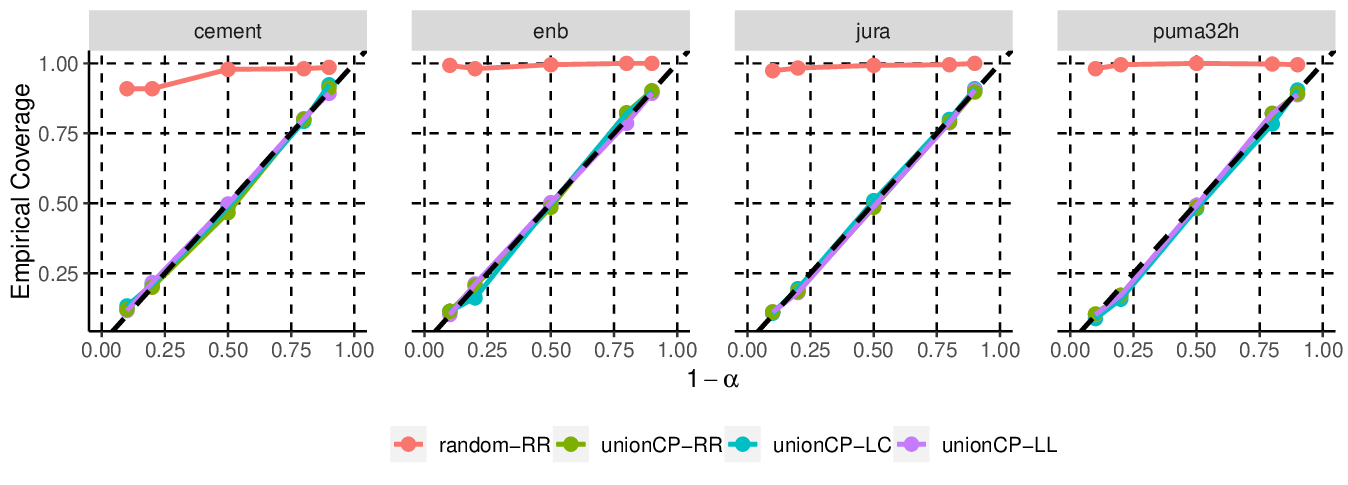}
    }
    \caption{Comparison of empirical coverage with random selection of $k$ regions and \texttt{unionCP} using ridge regression (RR), local-constant regression (LC) and local-linear regression (LL) across 10 repetitions.}
    \label{fig:compare-cp2}
\end{figure*}

Figure \ref{fig:compare-cp} includes a comparison of the $p$-value change-point regions constructed with \Cref{eqn:multi-conf} to the conformal prediction sets constructed using \texttt{gridCP}, both of which utilize linear regression for each predictor. Figure \ref{fig:compare-cp2} includes empirical coverage results for the approximation approaches described in Section \ref{sec:approx}, specifically with regions constructed using $||\cdot||^2_W$ in conjunction with ridge regression (RR), local constant regression (LC) and local-linear regression (LL). 

We also include results for the root-based approximation results described in Section \ref{sec:root-approx} for a wide-array of predictors that include those more general than the restrictions we outline in our paper. We emphasize based on the results in Figure \ref{fig:compare-cp2}, that \texttt{unionCP} results in consistently valid, but not overly conservative prediction regions. We omit volume comparison given the complex nature of the prediction sets constructed using \texttt{unionCP}.\looseness=-1


\subsection{\texttt{rootCP} Approximation Application}

We numerically examine the performance of \verb|rootCP| on multi-task regression problems using both synthetic and real databases. The experiments were conducted with a coverage level of $0.9$, \ie $\alpha = 0.1$.

For comparisons, we run the evaluations on $30$ repetitions of examples, and display the average of the following performance statistics for different methods: 1) the empirical coverage, \ie the percentage of times the prediction set contains the held-out target $y_{n+1}$, 2) the volume of the confidence intervals, and 3) the execution time. For each run, we randomly select a data tuple $(x_i, y_i)$ to constitute the targeted variables for which we will compute the conformal prediction set. The rest is considered as observed data $\Data_n$. Similar experimental settings are considered in \cite{lei2019fast}.\looseness=-1

  \begin{figure}
    \centering
    \subfigure[(Linear) Kernel Ridge (jura)]{\includegraphics[width=0.3\columnwidth]{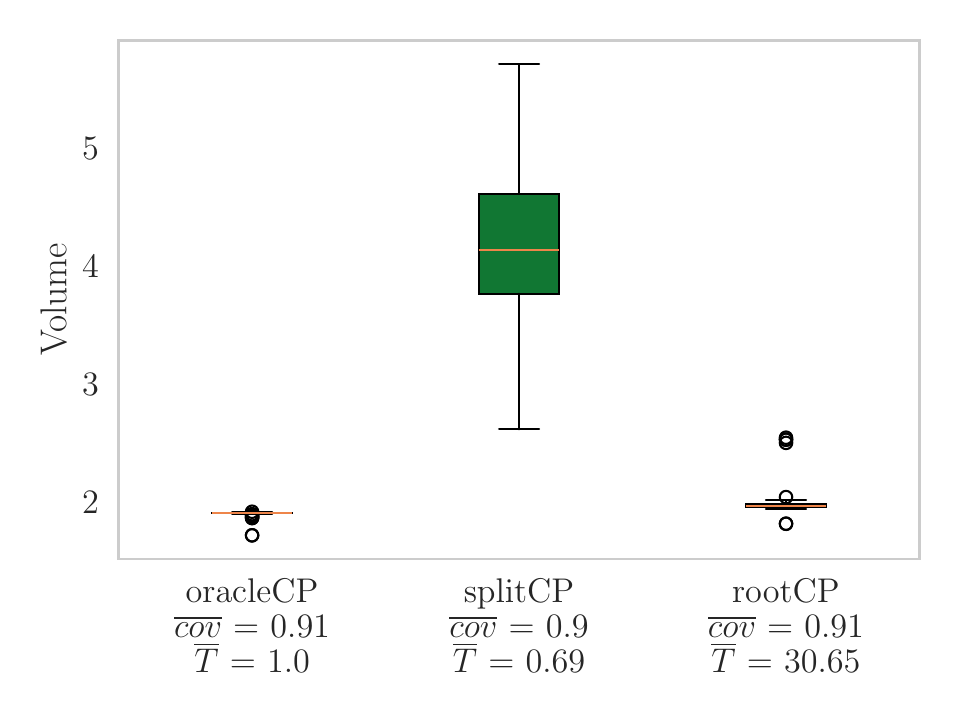}}
    \subfigure[(RBF) Kernel Ridge (enb)]{\includegraphics[width=0.3\columnwidth]{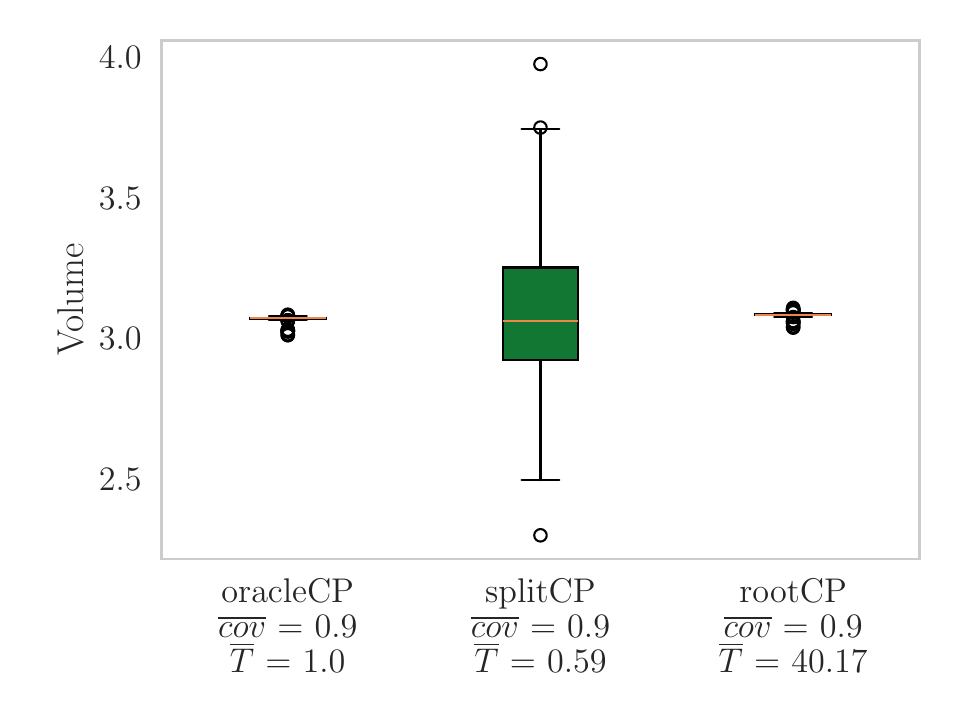}}
    \subfigure[Support Vector Regression (cement)]{\includegraphics[width=0.3\columnwidth]{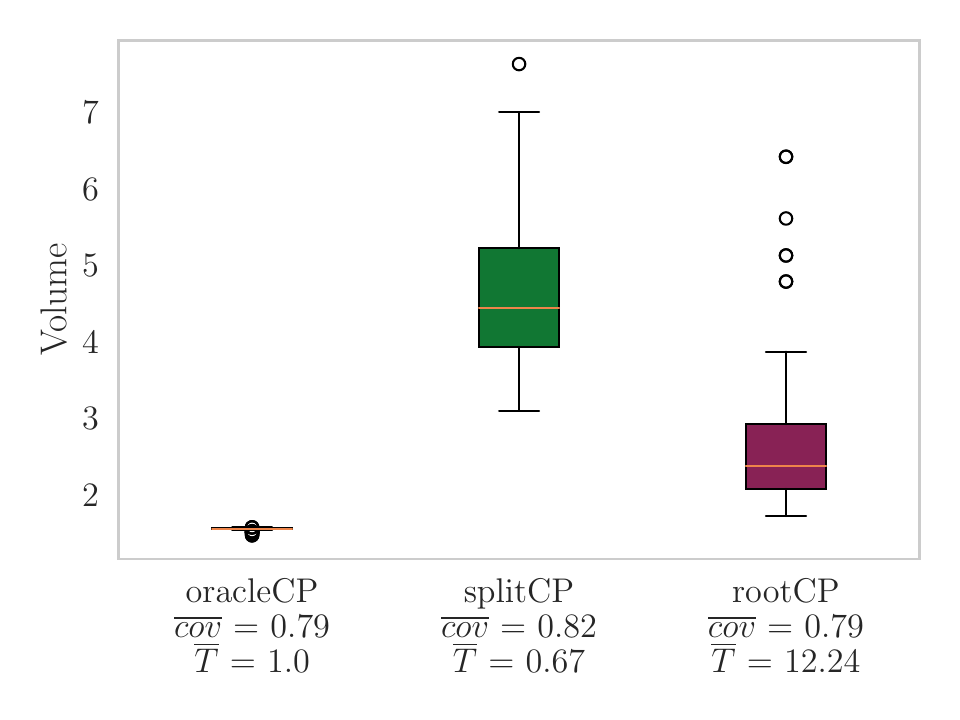}}
    \caption{Benchmarking the ellipse based conformal sets for several regression models on \texttt{jura}, \texttt{enb} and \texttt{cement} datasets. We display the volumes of the confidence sets over $100$ random permutation of the data. We denoted $\overline{cov}$ the average coverage, and $\overline{T}$ the average computational time normalized with the average time for computing \texttt{oracleCP} which requires a single model fit on the whole data. \label{fig:root_benchmarks2}\looseness=-1}
        \label{fig:rootcp_results}
\end{figure}
 
We run experiments on a suite of complex regression models, including: multi-task elastic net, multi-layer perceptron, orthogonal matching pursuit, kernel ridge regression with both linear and Gaussian kernels, support vector regression, $k$-nearest neighbors and quantile regression. A subset of results for some fo the real-world datasets are shown in \Cref{fig:rootcp_results}. We include additional results in Supplementary Materials. 

\section{Conclusion}
\label{sec:conclusion}

In this paper, we introduced exact $p$-values in multiple dimensions for predictors that are a linear function of the candidate value. Specifically, we discussed the exact construction of $p$-values using various conformity measures, including $\ell_1$ and $||\cdot||^2_W$. Additionally, we introduced methods for various approximations of multidimensional $1-\alpha$ conformal prediction sets through union-based and root-based prediction set construction, \verb|unionCP| and and a multi-task extension to \verb|rootCP|, respectively. We also also deliver probabilistic bounds and convergence results for these approximations. We then showed empirically with multiple predictors, including a subset of both linear and nonlinear predictors, that these approximations are comparable to \texttt{gridCP} sets, while drastically reducing the computational requirements. 



Other questions about the theoretical guarantees of our approach have yet to be answered. For example, we lack precise characterizations on the number of points to be sampled on the conformal set boundary, as well as implications our convex approximations, \eg ellipse, convex hull, related to expected volume and potential coverage loss in the worst case. Besides the conformal sets presented in this paper, these questions are equally relevant to the construction of any high-dimensional confidence sets. \looseness=-1

\section{Statements and Declarations}

\paragraph{Funding}
Partial financial support of CJ was received from the Air Force Office of Scientific Research.

\paragraph{Conflicts of Interest}
The authors have no competing interests to declare that are relevant to the content of this article.

\paragraph{Ethics Approval}
Not applicable.

\paragraph{Consent to Participate}
Not applicable.

\paragraph{Consent for Publication}
All contributors to the manuscript consent to publication. The views expressed in this article are those of the authors and do not reflect the official policy or position of the United States Air Force, United States Department of Defense, or United States Government.

\paragraph{Availability of Data and Material}
The real-world data utilized in our paper is publicly available and cited accordingly. Simulated data will be made public following acceptance.

\paragraph{Code Availability}
Supporting code will be made public following acceptance. 

\paragraph{Authors' Contributions}
Both CJ and EN provided equal contribution for this work, to include all conceptualization, formal analysis, writing, review and editing.

\bibliography{main.bib}
\bibliographystyle{apalike}
\clearpage


\end{document}